\documentclass{article}

    \usepackage[nonatbib, preprint]{neurips_2022}

\usepackage[utf8]{inputenc} % allow utf-8 input
\usepackage[T1]{fontenc}    % use 8-bit T1 fonts
\usepackage{hyperref}       % hyperlinks
\usepackage{url}            % simple URL typesetting
\usepackage{booktabs}       % professional-quality tables
\usepackage{amsfonts}       % blackboard math symbols
\usepackage{nicefrac}       % compact symbols for 1/2, etc.
\usepackage{microtype}      % microtypography
\usepackage{xcolor}         % colors
\usepackage{color, colortbl}
\usepackage{array,multirow}
\usepackage{amsmath, mathtools, bm, bbm}
\usepackage{amsthm}
\usepackage{amssymb}
\usepackage{tabularx}
\usepackage{wrapfig}
\usepackage{xspace}

\definecolor{nearwhite}{HTML}{FEFEFE}

\doublerulesepcolor{nearwhite}

\newcommand{\ie}{\textit{i}.\textit{e}., }
\newcommand{\eg}{\textit{e}.\textit{g}., }

\newcommand{\norm}[1]{\left\lVert#1\right\rVert}

\newcommand{\support}{\mathbb{S}}
\newcommand{\query}{\mathbb{Q}}

\newcommand{\cseen}{\mathbb{C}_{\text{CS}}}

\newcommand{\cunseen}{\mathbb{C}_{\text{OS}}}

\newcommand{\inputspace}{\mathcal{X}}

\newcommand{\featurespace}{\mathcal{Z}}
\newcommand{\dphiteta}{\mathcal{D}_{\phi_{\theta}}}

\newcommand{\OSTIM}{\textsc{OSTIM}\xspace}
\newcommand{\knn}{$k$-\textsc{NN}}

\newcommand{\zbf}{\boldsymbol{z}}
\newcommand{\xbf}{\boldsymbol{x}}

\newcommand{\lbf}{\boldsymbol{l}}
\newcommand{\pbf}{\boldsymbol{p}}
\newcommand{\mubf}{\boldsymbol{\mu}}
\newcommand{\wbf}{\boldsymbol{w}}
\newcommand{\thetabf}{\boldsymbol{\theta}}

\newcommand{\dotprod}[2]{\langle #1,~ #2 \rangle}

\definecolor{lightgray}{gray}{0.55}
\definecolor{atomictangerine}{rgb}{1.0, 0.6, 0.4}
\definecolor{carrotorange}{rgb}{0.93, 0.57, 0.13}
\definecolor{lightsalmon}{rgb}{1.0, 0.63, 0.48}

\newcommand{\std}[1]{\color{lightgray}{\scriptsize{$\pm{#1}$}}}
\newcommand{\magicpar}[1]{\smallskip\noindent {\textbf{#1}}\enskip}

\title{Model-Agnostic Few-Shot Open-Set Recognition}

\author{%
  Malik Boudiaf\hspace{0.15em}$^{\ast}$\\
  \'ETS Montreal
  \And
  Etienne Bennequin\thanks{Equal contribution. Corresponding authors: \{malik.boudiaf.1@etsmtl.net, etienneb@sicara.com\}} \\
  CentraleSupelec \\ Université Paris-Saclay \\ Sicara
  \And
  Myriam Tami \\
  CentraleSupélec \\ Université Paris-Saclay
  \And
  Celine Hudelot \\
  CentraleSupélec \\ Université Paris-Saclay
  \And
  Antoine Toubhans \\
  Sicara
  \And
  Pablo Piantanida \\
  CentraleSupélec-CNRS \\ Université Paris-Saclay  
  \And
  Ismail Ben Ayed \\
  \'ETS Montreal \\
}

\begin{document}

\maketitle

\begin{abstract}
    We tackle the Few-Shot Open-Set Recognition (FSOSR) problem, \ie classifying instances among a set of classes for which we only have few labeled samples, while simultaneously detecting instances that do not belong to any known class. Departing from existing literature, we focus on developing model-agnostic inference methods that can be plugged into any existing model, regardless of its architecture or its training procedure. Through evaluating the embedding's quality of a variety of models, we quantify the intrinsic difficulty of model-agnostic FSOSR. Furthermore, a fair empirical evaluation suggests that the naive combination of a \knn~detector and a prototypical classifier ranks before specialized or complex methods in the inductive setting of FSOSR. These observations motivated us to resort to \emph{transduction}, as a popular and practical relaxation of standard few-shot learning problems. We introduce an Open Set Transductive Information Maximization method (\OSTIM), which hallucinates an outlier prototype while maximizing the mutual information between extracted features and assignments. Through extensive experiments spanning 5 datasets, we show that \OSTIM surpasses both inductive and existing transductive methods in detecting open-set instances while competing with the strongest transductive methods in classifying closed-set instances. We further show that \OSTIM's model agnosticity allows it to successfully leverage the strong expressive abilities of the latest architectures and training strategies without any hyperparameter modification, a promising sign that architectural advances to come will continue to positively impact \OSTIM's performances. \footnote{Code is available at \url{https://github.com/ebennequin/few-shot-open-set}}.
\end{abstract}

\section{Introduction}
    Few-Shot Classification (FSC) consists in recognizing concepts, or classes, for which we only have a handful of labeled examples. These examples form the \textit{support set}. Most methods classify query instances based on their similarity to support instances in a feature space \cite{snell2017prototypical}, therefore implicitly assuming a \textit{closed-set} setting, in which query instances are supposed to be constrained to the set of explicitly defined classes. However, the real world is open. This closed-set assumption rarely holds in practice, especially for limited support sets. Whether they are unexpected items circulating on an assembly line, a new dress not yet included in a marketplace's catalog, or a previously undiscovered species of fungi, \textit{open-set instances} occur everywhere. When they do, a closed-set classifier will falsely label them as the closest known class.
    
    This drove the research community toward \textit{Open-Set Recognition} \ie recognizing instances with the awareness that they may belong to an unknown class. Although specifically designed methods in large-scale settings allow reliable detection of open-set instances while maintaining good accuracy on closed-set instances \cite{scheirer2012toward, bendale2016towards, zhou2021learning}, recent attempts at developing Few-Shot Open-Set Classifiers expose it to be a very difficult task \cite{liu2020few, jeong2021few}. Furthermore, these few-shot methods require a specific training strategy usually referred to as \textit{episodic training}. As a result, they cannot be applied to a new architecture without a tedious optimization of the training process.  We, on the other hand, argue that model-agnosticity, \ie the ability to be directly plugged into any trained model should be considered an important feature for a few-shot method, because 1) such a method can be seamlessly integrated into an existing pipeline without the need to re-train the model; 2) it can scale up to the latest and most significant advances in representation learning (\eg ViTs or self-supervised learning) without any additional effort; and 3) the difficult reproduction of episodic training has been shown to be an obstacle to the fair comparison between methods \cite{antoniou2018train, bennequin2019meta}, while this is not an issue when comparing methods using the same trained parameters. Therefore, in this work, we consider the challenge of building a method that acts on the feature space \ie the output space of an already trained model.
    
    We empirically observe that in this feature space, the clusters corresponding to the images' classes are far better defined for classes that were seen during the model's training than they are for novel classes, regardless of the model's architecture or training process. This has a direct and disastrous effect on FSOSR models: not only are they limited to a few labeled samples, but they must also perform outlier recognition on ill-defined class distributions. Our conclusions on the difficulty of FSOSR motivate us to consider a loose relaxation of this problem. In particular, the transductive setting is very popular in closed-set few-shot learning \cite{veilleux2021realistic, dhillon2019baseline, liu2020prototype, ziko2020laplacian, boudiaf2020transductive, wang2020instance, hu2021leveraging, boudiaf2021few} and has led to state-of-the-art performances by assuming access to the whole query set at once, therefore using it as additional unlabeled data. A limiting factor of existing transductive methods is that they implicitly assume that all query instances belong to the classes represented in the support set. When this is not the case, we show that transductive methods tend to artificially match the prediction confidence of closed-set and open-set instances, making prediction-based detection more difficult. To address this issue and fully exploit the potential of transduction in FSOSR, we introduce a novel Open Set Transductive Information Maximization method (\OSTIM), which builds on the InfoMax principle used in \cite{boudiaf2020transductive} and relaxes the closed-set assumption by hallucinating an outlier category.

    We show on both standard and exploratory Few-Shot Classification benchmarks that \OSTIM significantly surpasses its inductive and transductive predecessors alike for outlier detection while competing with the best transductive methods in terms of closed-set accuracy. Furthermore, unlike other FSOSR methods, and as motivated earlier, \OSTIM is an inference-only method. Applied on a wide variety of architectures and training strategies and without any re-optimization of its parameters, \OSTIM's improvement over a strong baseline is large and consistent, showing that our method can fully benefit from the latest advances in standard image recognition. Before diving into the core content, let us summarize our contributions:
    \begin{enumerate}
        % \item We study the differences between the distributions of classes' instances, depending on whether the classes have been seen during the model's training. To this end, we introduce the Mean Imposture Factor, a metric to measure how much the classes' distributions in a dataset are perturbed by instances from other classes. We expose the difficulty of detecting open-set instances among other classes, on a wide range of benchmarks and architectures.
        \item We expose the specific difficulty of the FSOSR problem when using off-the-shelf pre-trained models, on a wide range of benchmarks and architectures. To quantify this difficulty, we introduce the Mean Imposture Factor, a metric to measure how much the classes' distributions in a dataset are perturbed by instances from other classes. 
        % We expose the difficulty of detecting open-set instances among other classes, on a wide range of benchmarks and architectures.
        \item Stemming from the difficulty of the FSOSR problem, we introduce the transductive FSOSR setting, as a loose relaxation of the inductive FSOSR. We reproduce and benchmark five state-of-the-art model-agnostic transductive methods, and identify a shared issue. As a solution, we introduce \OSTIM for \textit{Open Set Transductive Information Maximization}.
        \item  Through extensive experiments spanning five datasets and a dozen of pre-trained models, we show that \OSTIM respects the model-agnosticity property, in that it consistently surpasses both inductive and existing transductive methods in detecting open-set instances while competing with the strongest transductive methods in classifying closed-set instances.
    \end{enumerate}

\section{Related Works}
    
\magicpar{Few-Shot Classification (FSC).}
    Many proposed methods to solve Few-Shot Classification involve episodic training \cite{Vinyals16}, in which a neural network acting as a feature extractor is trained on artificial FSC tasks sampled from the training set. This replication of the inference scenario during training is intended to make the learned representation more robust to new classes. However, recent works show that simple fine-tuning baselines are competitive in comparison to sophisticated episodic methods \cite{Chen19, goldblum2020unraveling}, motivating a new direction of Few-Shot Learning research towards the development of model-agnostic methods that do not involve any specific training strategy \cite{dhillon2019baseline}. 
    
\magicpar{Transductive FSC.}
    Transductive Few-Shot Classification methods leverage the query set as unlabeled data to improve classification, through model fine-tuning \cite{dhillon2019baseline}, Laplacian regularization \cite{ziko2020laplacian}, clustering \cite{lichtenstein2020tafssl}, mutual information maximization \cite{boudiaf2020transductive, veilleux2021realistic}, prototype rectification \cite{liu2020prototype}, or Optimal Transport \cite{bennequin2021bridging, hu2021leveraging, lazarou2021iterative}. As shown in Sec. \ref{sec:experiments}, these methods perform well on closed-set accuracy but fail at outlier detection. In comparison, our method leverages transduction to improve both aspects.

\magicpar{Open-Set Recognition (OSR).}
    OSR aims to enable classifiers to detect instances from unknown classes \cite{scheirer2012toward}. Prior works address this problem in the large-scale setting by augmenting the SoftMax activation to account for the possibility of unseen classes \cite{bendale2016towards}, generating artificial outliers \cite{ge2017generative, neal2018open}, improving closed-set accuracy \cite{vaze2021open}, or using placeholders to anticipate novel classes' distributions with adaptive decision boundaries \cite{zhou2021learning}.
    All these methods involve the training of deep neural networks on a specific class set and do not fully fit for the few-shot setting. In this work, we propose simple yet effective adaptations of OpenMax \cite{bendale2016towards} and PROSER \cite{zhou2021learning} as strong baselines for FSOSR.

\magicpar{Few-Shot Open-Set Recognition.}
    % Recent works develop novel solutions specific to the problem of Open-Set Recognition in a 
    In the few-shot setting, methods must detect open-set instances while only a few closed-set instances are available.
    % In FSOSR, the query set may contain images of classes that do not appear in the support set. 
    \cite{liu2020few} use meta-learning on pseudo-open-set tasks to train a model to maximize the classification entropy of open-set instances. \cite{jeong2021few} use transformation consistency to measure the divergence between a query image and the set of class prototypes. Both these methods require the optimization of a separate model with a specific episodic training strategy. Nonetheless, as we show in Section \ref{sec:experiments}, they bring marginal improvement over simple adaptations of standard OSR methods to the few-shot setting. In comparison, our method doesn't require any specific training and can be plugged into any feature extractor without further optimization.

\section{Few-Shot Open-Set Recognition} \label{sec:fsosr_setting}

    \magicpar{Setup and formalization.} 
        Given an input space $\inputspace$ and a \textit{closed-set} of classes $\cseen$, with $|\cseen|=K$, a K-way FSOSR task is formed by a support set of labeled instances $\support = \{(\xbf^s_i,y^s_i) \in \inputspace \times \cseen \}_{i=1...|\support|}$ and a query set $\query = \{\xbf^q_i \in \inputspace \}_{i=1...|\query|}$. In the standard few-shot setting, the unknown ground-truth query labels $\{y^q_i\}_{i=1...|\query|}$ are assumed to be restricted to closed-set classes \ie $\forall i, ~y^q_i \in \cseen$. In FSOSR, however, query labels may also belong  to an additional set $\cunseen$ of \textit{open-set} classes \ie $\forall i, ~y^q_i \in \cseen \cup \cunseen$ with $\cseen \cap \cunseen = \emptyset$.
        % from a query set $\query = \{(\xbf^q_i, y^q_i)\}_{i=1...|\query|}$. Unlike the standard Few-Shot setting, the ground truth label $y^q_i$ of query instance $i$ is not restricted to the \textit{closed-set} of classes $\cseen$, and can also belong to an additional set $\cunseen$ of \textit{open-set} classes, with $\cseen \cap \cunseen = \emptyset$. 
        For easy referencing, we refer to query samples from the closed-set classes $\cseen$ as \textit{inliers} and to query samples from open-set classes $\cunseen$ as \textit{outliers}. 
        % For each query image $\xbf_i^q$, the goal of FSOSR is to simultaneously assign a closed-set prediction $p_{ik} = p(y_i^q = k | \xbf_i^q),~ k \in \cseen$  and an \textit{outlierness} score $p(y^q_i \not \in \cseen | \xbf_i^q)$. 
        For each query image $\xbf_i^q$, the goal of FSOSR is to simultaneously assign a closed-set prediction $p^q_{ik} = \mathbbm P(y_i^q = k | \xbf_i^q),~ k \in \cseen$  and an \textit{outlierness} score $\mathbbm P(y^q_i \not \in \cseen | \xbf_i^q)$. 
        As part of the model-agnostic setting, we perform this prediction on top of features lying in a feature space $\featurespace$ and extracted by a frozen model $\phi_{\thetabf}: \inputspace \rightarrow \featurespace$, whose parameters $\thetabf$ were trained on some large dataset $\mathcal{D}_{base} = \{(\xbf^b_i,y^b_i)\}_{i=1...|\mathcal{D}_{base}|}$ such that for all $i$, $y^b_i \in \mathbb{C}_{base}$ with $\mathbb{C}_{base} \cap \cseen = \mathbb{C}_{base} \cap \cunseen = \emptyset$.

        % An $K$-way $N_s$-shot few-shot classification task is such that $|\cseen| = K$ and the support set contains $N_s$ labelled instances for each class, typically $N_s<10$. 
        
        \begin{wrapfigure}{r}{0.5\textwidth}
            \centering
            \vspace{-14pt}
            \includegraphics[width=.24\textwidth]{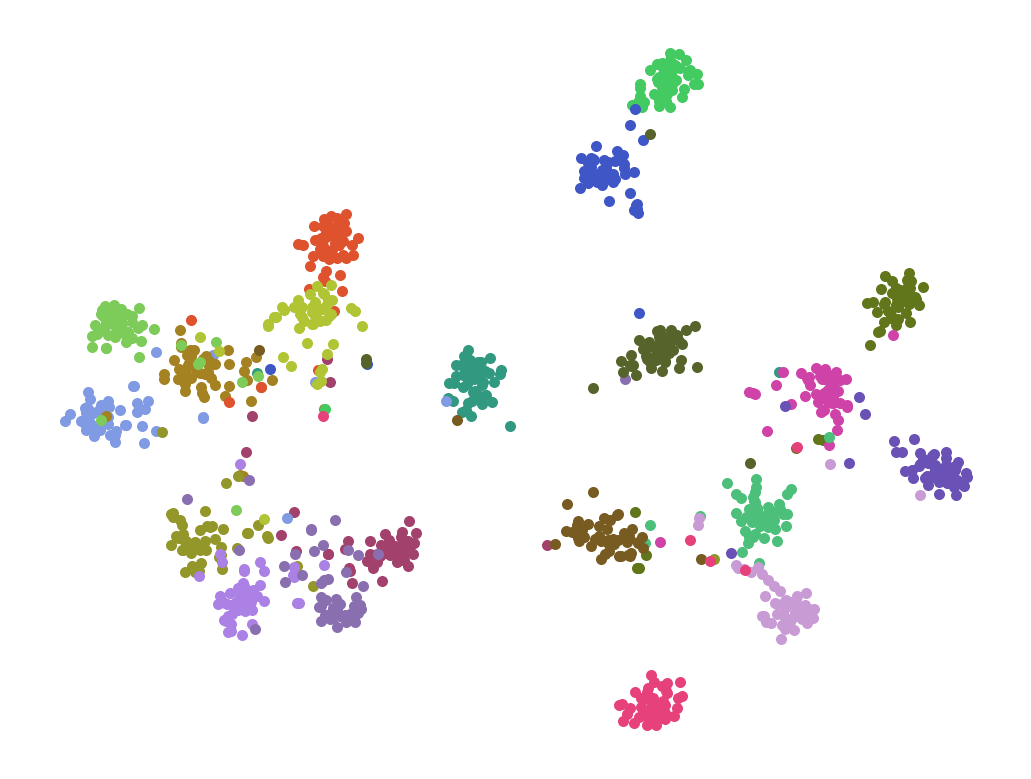}
            \includegraphics[width=.24\textwidth]{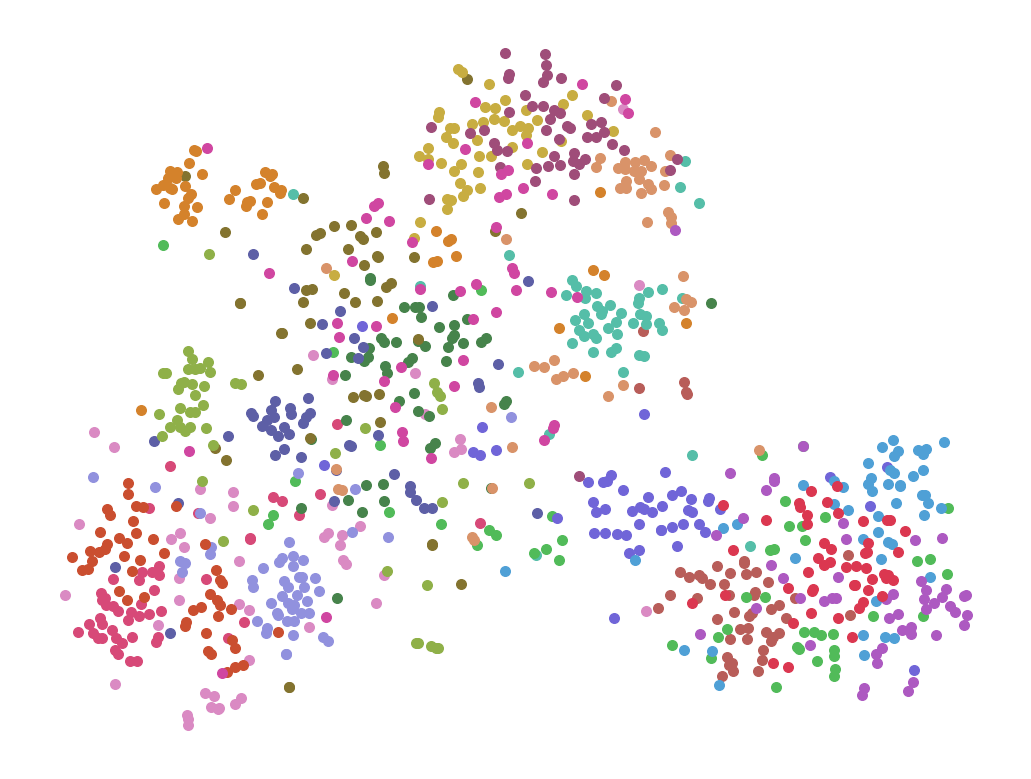}
            \caption{2-dimensional reduction with T-SNE of feature extracted from ImageNet's validation set using a ResNet12 trained on \textit{mini}ImageNet. (Left): images from 20 randomly selected classes represented in \textit{mini}ImageNet's base set.  (Right): Images from the 20 classes represented in \textit{mini}ImageNet's test set. Each color corresponds to a distinct class.}
            \label{fig:clustering-mini-imagenet}
            \vspace{-10pt}
        \end{wrapfigure}

    \magicpar{Measuring the difficulty of outlier detection on novel classes.} \label{subsec:difficulty_of_problem}
        As an anomaly detection problem, open-set recognition consists in detecting samples that differ from the population that is known by the classification model. However, in FSOSR, neither closed-set classes nor open-set classes have been seen during the training of the feature extractor \ie $\mathbb{C}_{base} \cap \cseen =  \mathbb{C}_{base} \cap \cunseen = \emptyset$. In that sense, both the inliers and the outliers of our problem can be considered outliers from the perspective of the feature extractor. Intuitively, this makes it harder to detect open-set instances, since the model doesn't know well the distribution from which they are supposed to diverge. Here we empirically demonstrate and quantify the difficulty of OSR in a setting where closed-set classes have not been represented in the training set. Specifically, we estimate the gap in terms of quality of the classes' definition in the feature space, between classes that were represented during the training of the feature extractor \ie $\mathbb{C}_{base}$, and the classes of the test set, which were not represented in the training set. To do so, we introduce the novel Mean Imposture Factor measure and use the intra-class to inter-class variance ratio $\rho$ as a complementary measure. Note that the following study is performed on whole datasets, \textit{not} few-shot tasks.
            
        \magicpar{Mean Imposture Factor (MIF).}Let $\dphiteta \subset \featurespace \times \mathbb C$ be a labeled dataset of extracted feature vectors, with $\phi_{\thetabf}$ a fixed feature extractor and $\mathbb C$ a finite set of classes. For any feature vector $\zbf$ and a class $k$ to which $\zbf$ does not belong, we define the Imposture Factor $\textit{IF}_{\zbf|k}$ as the proportion of the instances of class $k$ in $\dphiteta$ that are further than $\zbf$ from their class centroid. Then the MIF is the average IF over all instances in $\dphiteta$.
        % In a labeled dataset of extracted feature vectors $\dphiteta \subset \featurespace \times \mathbb C$ with $\phi_{\thetabf}$ a fixed feature extractor and $\mathbb C$ a finite set of classes, the MIF is the average Imposture Factor (IF) over all instances in $\mathcal{\dphiteta}$ and classes in $\mathbb C$. We define the IF for any feature vector $\zbf$ with respect to any class $k \in \mathbb C$ to which $\zbf$ doesn't belong as the proportion of instances with class $k$ which are further than $\zbf$ from their class centroid $\mubf_k$ \ie
            \begin{equation}
            \boxed{
                \textit{MIF} = \frac{1}{|\mathbb C|} \sum_k \frac{1}{|\dphiteta \backslash \mathcal D_k|} \sum_{\zbf \notin \mathcal D_k} \textit{IF}_{\zbf|k}
            }
                ~\textrm{ with  }~
                \textit{IF}_{\zbf|k} = \frac{1}{|\mathcal D_k|} \sum_{\zbf' \in \mathcal D_k} \mathbbm{1}_{\|\zbf'-\mubf_k\|_2>\|\zbf-\mubf_k\|_2}
                % \left[ \|\zbf'-\mubf_k\|_2>\|\zbf-\mubf_k\|_2 \right]
            \end{equation}
        with  $\mathcal D_k$ the set of instances in $\dphiteta$ with label $k$, and $\mathbbm{1}$ the indicator function. The MIF is a measure of how perturbed the clusters corresponding to the ground truth classes are. A MIF of zero means that all instances are closer to their class centroid than any outsider. Note that $\text{MIF} = 1 - \text{AUROC}(\psi)$ where $\text{AUROC}(\psi)$ is the area under the ROC curve for an outlier detector $\psi$ that would assign to each instance an outlier score equal to the distance to the ground truth class centroid. To the best of our knowledge, the MIF is the first tool allowing to measure the class-wise integrity of a projection in the feature space. As a sanity check for MIF, we also report the intra-class to inter-class variance ratio $\rho$, used in previous works \cite{goldblum2020unraveling}, to measure the compactness of a clustering solution.
        
        % ratio between the average square distance of all instances to their respective class centroid and the average square distance of class centroids to the average feature vector of the whole dataset. This  to measure the quality of a dataset's projection in a feature space.
        
        % we also resort to the ratio between the average square distance of all instances to their respective class centroid and the average square distance of class centroids to the average feature vector of the whole dataset. This intra-class to inter-class variance ratio $\rho$ is used in previous works \cite{goldblum2020unraveling} to measure the quality of a dataset's projection in a feature space.
        
        \magicpar{Base classes are better defined than test classes.} We experiment on three widely used Few-Shot Learning benchmarks: \textit{mini}ImageNet \cite{Vinyals16}, \textit{tiered}ImageNet \cite{tiered_imagenet}, and ImageNet $\rightarrow$ Aircraft \cite{maji2013fine}. We use the validation set of ImageNet in order to obtain novel instances for ImageNet, \textit{mini}ImageNet, and \textit{tiered}ImageNet's base classes. We also use it for test classes for consistency.
        In Figure \ref{fig:clustering-mini-imagenet}, we present a visualization of the ability of a ResNet12 trained on \textit{mini}ImageNet to project images of both base and test classes into clusters. While we are able to obtain well-separated clusters for base classes after the 2-dimensional T-SNE reduction, this is clearly not the case for test classes, which are more scattered and overlapping.  Such results are quantitatively corroborated by Table \ref{tab:clustering-results}, which shows that both \text{MIF} and $\rho$ are systematically lower for base classes across 3 benchmarks and 5 feature extractors.
        This demonstrates the difficulty of defining in the feature space the distribution of a class that was not seen during the training of the feature extractor, and therefore the difficulty of defining clear boundaries between inliers and outliers \ie closed-set images and open-set images, all the more when only a few samples are available. 
        
        \begin{table}[t]
            
            \centering
            \small
            \label{tab:clustering-results}
            \caption{Contrast between datasets made of images from classes represented (\textit{base}) or not represented (\textit{test}) in the feature extractor's training set, on three benchmarks and with several backbones (RN12: ResNet12, WRN: WideResNet1810,  ViT, RN50: ResNet50, and MX: MLP-Mixer), following the MIF (in percents) and the variance ratio ($\rho$). Best result for each column is shown in bold. }
             \resizebox{\textwidth}{!}{
            \begin{tabular}{lcccccccccccccc}
            
            \toprule
             \multirow{3}{*}{Classes} & \multicolumn{4}{c}{miniImageNet} & \multicolumn{4}{c}{tieredImageNet} & \multicolumn{6}{c}{ImageNet $\rightarrow$ Aircraft} \\
                  \cmidrule(rl){2-5} \cmidrule(lr){6-9} \cmidrule(lr){10-15}
                  & \multicolumn{2}{c}{$\rho$} & \multicolumn{2}{c}{MIF (\%)}      & \multicolumn{2}{c}{$\rho$} & \multicolumn{2}{c}{MIF (\%)}      & \multicolumn{3}{c}{$\rho$}            & \multicolumn{3}{c}{MIF (\%)}                      \\
                  \cmidrule(rl){2-3} \cmidrule(lr){4-5} \cmidrule(lr){6-7} \cmidrule(rl){8-9} \cmidrule(lr){10-12} \cmidrule(lr){13-15}
                  & RN12         & WRN         & RN12      & WRN       & RN12         & WRN         & RN12      & WRN       & ViT    & RN50          & MX           & ViT    & RN50          & MX           \\
                  \midrule
            \textit{base} & \textbf{0.93}    & \textbf{0.84}   & \textbf{0.89} & \textbf{1.03} & \textbf{1.09}    & \textbf{0.78}   & \textbf{0.78} & \textbf{0.81} & \textbf{0.96} & \textbf{1.36} & \textbf{2.54} & \textbf{0.09} & \textbf{0.29} & \textbf{0.31} \\
            \textit{test}  & 2.10             & 2.07            & 5.56          & 7.36          & 2.10             & 1.54            & 4.39          & 5.18          & 3.20          & 4.88          & 5.35          & 18.08          & 21.58          & 17.27         \\
            \bottomrule
            \end{tabular}
            }
            
        \end{table}

\section{Transductive FSOSR} \label{sec:transductive_ofsl}

	As a growing part of the Few-Shot literature, Transductive Few-Shot Learning assumes that unlabelled samples from the query set are observed at once, such that the structure of unlabelled data can be leveraged to help constrain ambiguous few-shot tasks. In practice, transductive methods have achieved impressive improvements over inductive methods in standard closed-set FSC \cite{dhillon2019baseline, boudiaf2020transductive, ziko2020laplacian, hu2021leveraging}. Considering the difficulty posed by the FSOSR problem, and detailed in Sec. \ref{subsec:difficulty_of_problem}, we expect that transductive methods can help us improve outlier detection while still achieving super-inductive closed-set predictive performance. Unfortunately, we empirically show in Sec. \ref{sec:experiments} that significant accuracy gains systematically come along with significant outlier detection degradation.  
	
	% In other words, existing transductive methods indeed tend to increase accuracy, but worsen calibration with respect to inductive methods. In the following paragraph, we illustrate the matter through the motivating example of the InfoMax transductive principle used in \cite{boudiaf2020transductive}, upon which our proposed method builds.
    	
    \begin{figure}
        \centering
        \begin{minipage}{.54\textwidth}
          \centering
         \includegraphics[width=.95\textwidth]{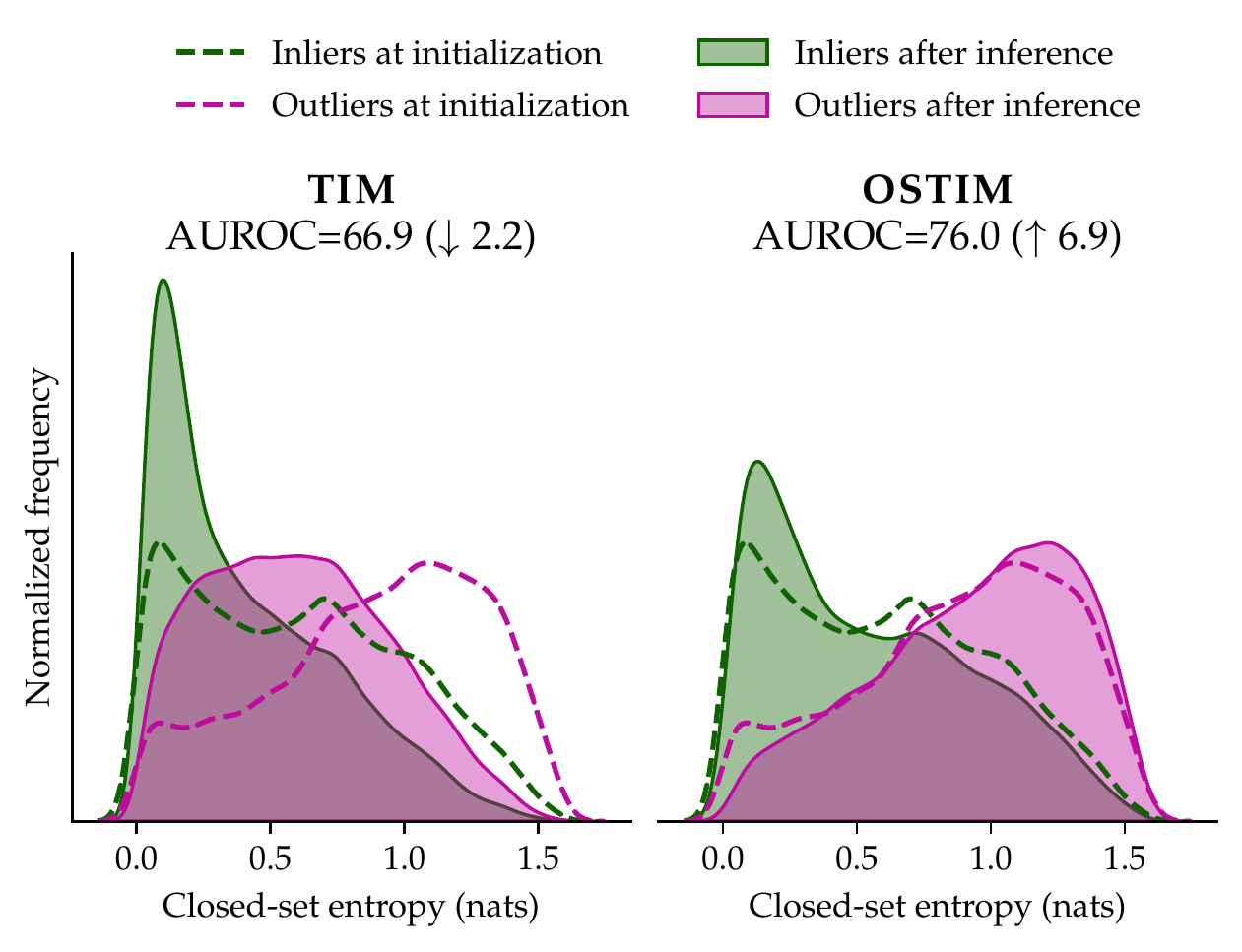}
          \caption{\textbf{Closed v.s. Open-Set InfoMax.} (Left) Minimizing closed-set entropy \cite{boudiaf2020transductive} on all samples degrades prediction-based outlier detection. (Right) \OSTIM tends to bin outliers in a $(K+1)^{th}$ category. Therefore, their \textit{open-set} conditional entropy in Eq. \eqref{eq:ottim_objective} decreases while their closed-set entropy increases.}
          \label{fig:entropy_histograms}
        \end{minipage}%
        \hspace{0.02\textwidth}
        \begin{minipage}{.43\textwidth}
          \centering
          \includegraphics[width=.87\textwidth]{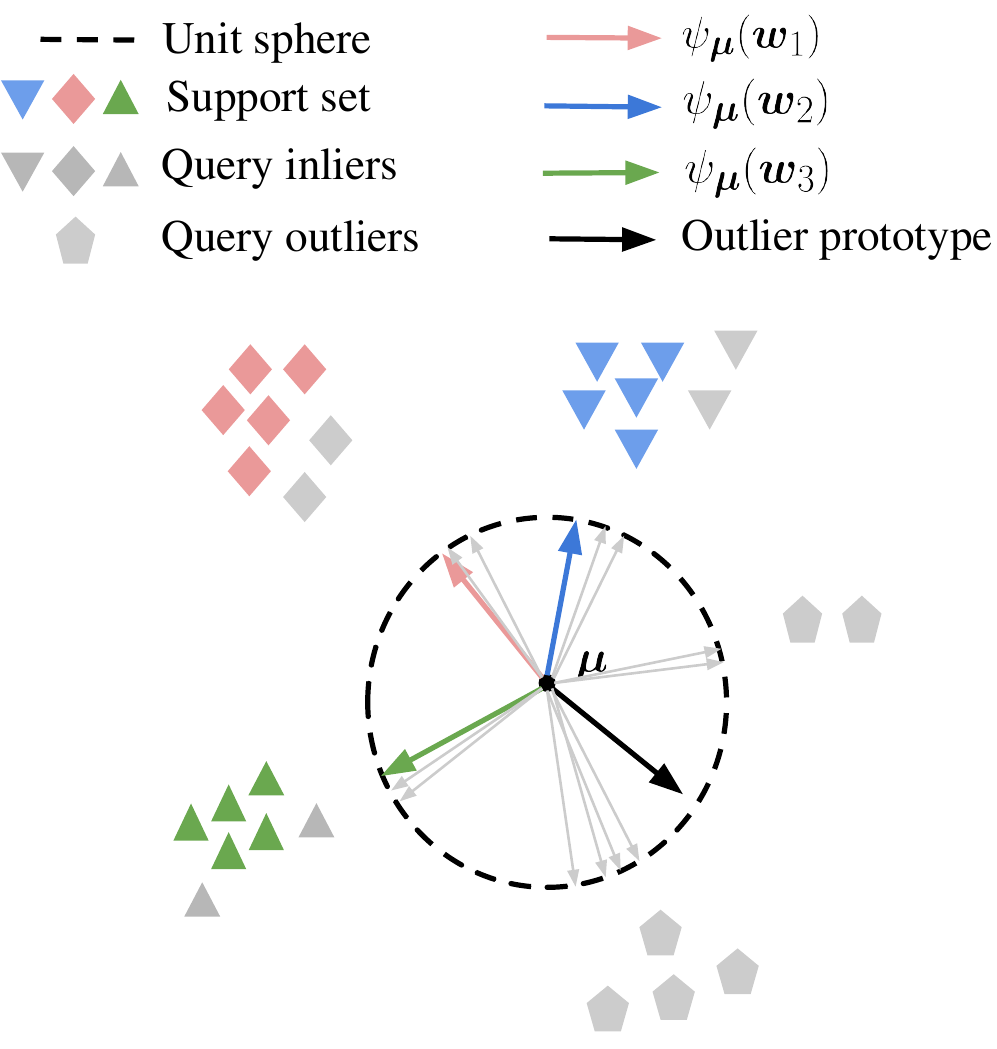}
          \caption{\textbf{Geometric intuition behind \OSTIM}. The \text{CE} term encourages colored arrows to align with support samples, while $\mathcal{I}_{\alpha}$ encourages grey arrows to either align with a colored arrow (inlier prototypes) or with the black arrow (outlier prototype).}
          \label{fig:geometric_intuition}
        \end{minipage}
    \end{figure}

	\magicpar{Diagnosing the InfoMax transduction.} Among the 5 transductive methods evaluated, we find TIM \cite{boudiaf2020transductive} offers the best trade-off between performances in closed-set classification and outlier detection, although the latter still falls far below inductive alternatives. Indeed, as part of the InfoMax principle, TIM \cite{boudiaf2020transductive} systematically enforces confident predictions on each query sample through conditional entropy minimization, whether this sample is an outlier or not. To make things worse, outliers' initial predictions typically fall in the region of the simplex where the magnitude of entropy's gradients is the highest \cite{veilleux2021realistic}, meaning the model prioritizes minimizing the entropy of outliers over inliers. Altogether, those ingredients lead to a degradation of the discriminability between inliers and outliers' predictions during inference. This situation is depicted in the left plot of Fig. \ref{fig:entropy_histograms}, where we observe the entropy histogram of outliers (purple) shifting significantly towards the left (low-entropy) after inference. Following these observations, we seek to instantiate the InfoMax principle in a way that simultaneously benefits closed-set predictive performance and outlier detection.

	\magicpar{Introducing OSTIM.} To help remediate the issue, we propose a simple yet highly effective modification to the original closed-set TIM \cite{boudiaf2020transductive} method, that retains TIM's high closed-set accuracy while drastically improving outlier detection. Importantly, it does not introduce any computational overhead or tunable hyperparameter. Relaxing the closed-set assumption, we consider a $K+1$-way classification problem, where the added class represents the broad \textit{outlier} category. We observe in Sec. \ref{sec:ablation_study} that introducing additional learnable parameters, e.g. a new prototype as in \cite{zhou2021learning} to represent the \textit{outlier} class in such low-data regimes yields poor performances. Consequently, we propose an implicit definition of the \textit{outlier} class that reuses existing parameters and remains differentiable. We name our method \OSTIM for \textit{Open Set Transductive Information Maximization}.

	\magicpar{Implicit outlier prototype.} Following the setting described in Sec. \ref{sec:fsosr_setting}, we abstract the base model and work directly on top of extracted features $\zbf=\phi_{\theta}(\xbf)$.
	We further consider the center-normalize transformation $\psi_{\mubf}: \mathcal{Z} \rightarrow \mathcal{Z}$, and define the similarity $l_{ik}$ between a sample $\zbf_i$ and a class prototype $\wbf_k$ as the dot-product between centered-normalized features and prototypes:
	\begin{align} \label{eq:inlier_logits}
		l_{ik} = \text{sim}({\zbf_i, \wbf_k}) = \dotprod{\psi_{\mubf}(\zbf_i)}{\psi_{\mubf}(\wbf_k)}, \quad \text{with} ~ \psi_{\mubf}(\zbf) = \frac{\zbf - \mubf}{~\norm{\zbf - \mubf}_2}.
	\end{align}
	Note that the original TIM uses standard L2-normalization of features, which amounts to fixing $\mubf=0$. SimpleShot \cite{wang2019simpleshot} fixes $\mubf$ as the mean of features from the base set, while PT-MAP \cite{hu2021leveraging} uses the mean of all features in the task. We adopt the latter in our main experiments, and ablate on this choice in Sec. \ref{sec:ablation_study}. We now define the \textit{outlier} logit as the negative average of inliers class logits:
	\begin{align} \label{eq:outlier_logit}
		l_{i,K+1} = - \frac{1}{K} \sum_{k=1}^{K} l_{ik} = \dotprod{\psi_{\mubf}(\zbf_i)}{\underbrace{- \frac{1}{K} \sum_{k=1}^{K} \psi_{\mubf}(\wbf_k)}_{\text{implicit \textit{outlier} prototype}}}.
	\end{align}
	The outlier logit can be interpreted as the similarity between some point and an \textit{outlier} prototype corresponding to the diametrical opposite of the average of inlier prototypes. To clarify this intuition, a geometrical description of the problem is provided in Figure \ref{fig:geometric_intuition}. For a center-normalized query point represented by a gray arrow, inlier logits $\{l_k\}_{k=1}^K$ correspond to measuring the angles between respective colored arrows and the gray arrow, while $l_{K+1}$ measures the similarity with the black arrow. The concatenation of \textit{inlier} logits and the \textit{outlier} logit forms the final logit vector $\lbf_i = [l_{i1}, \dots, l_{i,K+1}]^T$, which is translated into a probability vector $\pbf_i$ over the $K+1$ outcomes through a standard softmax operation. The first $K$ components of this probability vector $\{p_{ik}\}_{k=1}^K$ are used for closed-set classification, while the last $p_{i, K+1}$ is used as the outlierness score.

	\magicpar{Prototype refinement.} The prototypes $\{\wbf_k\}_{k=1}^K$ are initialized as the class-centroids using labeled samples from $\support$, and further refined by minimizing an open-set version of TIM's transductive loss:
% 	\begin{align}\label{eq:ottim_objective}
% 		\begin{split}
% 	    \boxed{\min_{\wbf} \ \textrm{CE} -   \mathcal{\widehat{I}}_{\alpha}} \
% 	    \textrm{ with } \ \textrm{CE} &\coloneqq -\frac{1}{|\support|} \sum_{i \in \support}\sum_{k=1}^{K+1} y_{ik} \log (p_{ik}), \\
% 	                 -  \mathcal{\widehat{I}}_{\alpha} &\coloneqq \underbrace{\sum_{k=1}^{K+1} \widehat{p}_{k} \log \widehat{p}_{k}}_{\substack{\text{marginal entropy} \\ \text{prevents trivial solutions}} }  \underbrace{- ~ \frac{\alpha}{|\query|} \sum_{i \in \query}\sum_{k=1}^{K+1} p_{ik} \log (p_{ik}), }_{\substack{\text{conditional entropy} \\ \text{forces query samples into} \\ \text{inlier category or outlier group}}}
% 	    \end{split}
% 	\end{align}
% 		where $y_{ik}=\mathbbm{1}[y_i = k], k \in [1, K]$ is a one-hot encoded version of the ground-truth label introduced in Sec. \ref{sec:fsosr_setting}, complemented with a last outlier component $y_{i,K+1}=0$, and $\hat{p}_k=\frac{1}{|\query|} \sum_i p_{ik}$ denotes the marginal prediction for class $k$.
		\begin{align}\label{eq:ottim_objective}
		\begin{split}
	    \boxed{\min_{\wbf} \ \textrm{CE} -   \mathcal{\widehat{I}}_{\alpha}} \
	    \textrm{ with } \ \textrm{CE} &\coloneqq -\frac{1}{|\support|} \sum_{i = 1}^{|\support|}\sum_{k=1}^{K+1} y^s_{ik} \log (p^s_{ik}), \\
	                 -  \mathcal{\widehat{I}}_{\alpha} &\coloneqq \underbrace{\sum_{k=1}^{K+1} \widehat{p}_{k} \log \widehat{p}_{k}}_{\substack{\text{marginal entropy} \\ \text{prevents trivial solutions}} }  \underbrace{- ~ \frac{\alpha}{|\query|} \sum_{i = 1}^{|\query|}\sum_{k=1}^{K+1} p^q_{ik} \log (p^q_{ik}), }_{\substack{\text{conditional entropy} \\ \text{forces query samples into} \\ \text{inlier category or outlier group}}}
	    \end{split}
	\end{align}
	where $y^s_{ik}=\mathbbm{1}[y^s_i = k], k \in [1, K]$ is a one-hot encoded version of the ground-truth label introduced in Sec. \ref{sec:fsosr_setting}, complemented with a last outlier component $y^s_{i,K+1}=0$, and $\hat{p}_k=\frac{1}{|\query|} \sum_i p^q_{ik}$ denotes the marginal prediction for class $k$. 
	Following \cite{boudiaf2020transductive}, $\alpha \in \mathbb R$ is found through validation. 
	Note that unlike in the standard TIM, the introduction of an additional $(K+1)^{th}$ class, represented by the implicit prototype, makes it possible to minimize the entropy of all samples without losing discriminability between inliers and outliers. In other words, outliers can simply be predicted in the  $(K+1)^{th}$ category with high confidence, and inliers in their associated inlier category. As a matter of fact, Fig. \ref{fig:entropy_histograms} shows that inliers' closed-set entropy decreases, indicating that they tend to get closer to some inlier prototype, while outliers' closed-set entropy increases, indicating that they are on average moving away from their closest inlier prototype. We emphasize that closed-set entropy is simply used for diagnosis, and neither corresponds to the outlierness score used by TIM \cite{boudiaf2020transductive} nor \OSTIM in Sec. \ref{sec:experiments}.

\section{Experiments} \label{sec:experiments}

    \subsection{Experimental setup}
    
        \begin{table}[t]
        \centering
        \small
        \caption{\textbf{Standard Benchmarking}. Evaluating different families of methods on the FSOSR problem, on the popular \textit{mini}-ImageNet, using a ResNet-12. For each column, a light-gray standard deviation is indicated, corresponding to the maximum deviation observed across methods for that metric. Best methods are shown in bold. Results for PEELER$^\star$ are reported from \cite{jeong2021few}. The same table for \textit{tiered}-ImageNet is available in Appendix, along with the details about the presented metrics.}
        \resizebox{\textwidth}{!}{
            \begin{tabular}{lccccccccc}
                \multicolumn{10}{c}{\textbf{\textit{mini}-ImageNet}} \vspace{0.5em}\\
                \toprule
                \multirow{3}{*}{Strategy} & \multirow{3}{*}{Method} & \multicolumn{4}{c}{1-shot} & \multicolumn{4}{c}{5-shot} \\
                \cmidrule(lr){3-6} \cmidrule(lr){7-10}
                & & Acc & AUROC & AUPR & Prec@0.9 & Acc & AUROC & AUPR & Prec@0.9 \\
                & & \std{0.72} & \std{0.79} & \std{0.69} & \std{0.47} & \std{0.44} & \std{0.73} & \std{0.61} & \std{0.56} \\
                \midrule
                \multirow{6}{*}{OOD detection} & \knn \cite{knn_detector} & - & 70.28 & 69.78 & 58.01 & - & 76.56 & 76.75 & 61.58 \\
                                               & IForest \cite{iforest_detector} & - & 56.10 & 55.54 & 52.34 & - & 63.46 & 62.21 & 55.09 \\
                                               & OCVSM \cite{ocsvm_detector} & - & 69.25 & 69.15 & 57.35 & - & 69.25 & 66.38 & 59.43 \\
                                               & PCA \cite{pca_detector} & - & 67.14 & 66.33 & 56.78 & - & 75.31 & 75.59 & 60.54 \\
                                               & COPOD \cite{copod_detector} & - & 50.45 & 51.76 & 50.86 & - & 52.70 & 53.6 & 51.47 \\
                                               & HBOS  & - & 58.25 & 57.39 & 53.32 & - & 61.62 & 60.64 & 54.47 \\

                \toprule
                \multirow{3}{*}{Inductive classifiers} & SimpleShot \cite{wang2019simpleshot} & 65.93 & 64.44 & 63.27 & 55.29 & 81.73 & 70.92 & 70.36 & 58.13 \\
                                                       & Baseline ++ \cite{Chen19} &  65.91 & 64.61 & 63.32 & 55.41 & 81.87 & 66.65 & 65.74 & 56.53 \\
                                                       & FEAT \cite{ye2020few} & 67.28 & 52.25 & 54.28 & 50.0 & 81.89 & 53.25 & 56.36 & 50.0 \\
                \midrule
                \multirow{4}{*}{Inductive Open-Set} & PEELER$^\star$ \cite{liu2020few} & 65.86 & 60.57 & - & - & 80.61 & 67.35 & - & - \\
                                                    & SnatcherF \cite{jeong2021few}  & 67.28 & 69.26 & 68.98 & 57.57 & 81.89 & 76.87 & 77.35 & 62.09  \\
                                                    & OpenMax \cite{bendale2016towards} & 65.93 & 70.86 & 70.34 & 58.30 &  82.37 & 77.34 & 77.62 & 62.14  \\
                                                    & PROSER \cite{zhou2021learning} & 64.89 & 68.63 & 68.42 & 57.19 & 80.09 & 75.0 & 75.64 & 60.18 \\
                \toprule
                \multirow{5}{*}{Transductive classifiers} & LaplacianShot \cite{ziko2020laplacian} & \textbf{70.99} & 51.88 & 53.59 &     51.66 &  \textbf{82.88} & 57.17 & 57.88 &     52.66   \\
                                                          & BDCSPN \cite{liu2020prototype} &  69.88 & 56.98 &     57.87 &     52.12 & \textbf{82.50}  & 61.52 &     62.29 &     53.44 \\
                                                          & TIM-GD \cite{boudiaf2020transductive} & 67.78 & 61.94 &     60.71 &     54.48 & \textbf{82.51} & 67.29 &     66.05 &     56.87 \\
                                                          & PT-MAP \cite{hu2021leveraging} & 66.29 & 58.66 &    58.32 &     53.44 & 77.72 & 63.22 &     62.89 &     54.77 \\
                                                          & LR-ICI \cite{wang2020instance} & 69.34 & 46.10	& 49.24 & 50.02 & 81.46 & 45.97 & 49.84 & 49.91 \\
                \midrule
                % \rowcolor{lightsalmon!25} Transductive Open-Set & \OSTIM (ours) & 69.23 & \textbf{73.08} & \textbf{74.32} &  \textbf{58.96} &  \textbf{82.79} & \textbf{83.80}  & \textbf{84.30}  &  \textbf{67.46} \\
                \rowcolor{lightsalmon!25} Transductive Open-Set & \OSTIM (ours) & 69.23 & \textbf{74.3} & \textbf{75.01} &  \textbf{59.93} &  \textbf{82.79} & \textbf{83.80}  & \textbf{84.30}  &  \textbf{67.46} \\
                \bottomrule
            \end{tabular}
        } 
        \label{tab:benchmark_results_mini}
    \end{table}    
        
        \magicpar{Baselines.} One goal of this work is to fairly evaluate different strategies to address the FSOSR problem. In particular, we benchmark 4 families of methods: (i) popular Outlier Detection methods, e.g. Nearest-Neighbor \cite{knn_detector}, (ii) Inductive Few-Shot classifiers, e.g. SimpleShot \cite{wang2019simpleshot} (iii) Inductive Open-Set methods formed by standard methods such as OpenMax \cite{bendale2016towards} and Few-Shot methods such as Snatcher \cite{jeong2021few} (iv) Transductive classifiers, e.g. TIM \cite{boudiaf2020transductive}, that implicitly rely on the closed-set assumption, and finally (v) Transductive Open-Set introduced in this work through \OSTIM. Following \cite{jeong2021few}, closed-set few-shot classifiers are turned into open-set classifiers by considering the negative of the maximum probability as a measure of outlierness. The center-normalize transformation $\psi_{\mubf}$ was found to benefit all methods. Therefore, we apply it to the features before applying any method, using an inductive \textit{Base centering} \cite{wang2019simpleshot} for inductive methods $\mubf_{Base}=\frac{1}{|\mathcal{D}_{base}|} \sum_{\xbf \in \mathcal{D}_{base}} \phi_{\thetabf}(x)$, and a transductive \textit{Task centering} \cite{hu2021leveraging} ${\mubf_{Task}=\frac{1}{|\support \cup \query|} \sum_{\xbf \in \support \cup \query} \phi_{\thetabf}(x)}$ for all transductive methods.

        \magicpar{Hyperparameters.} For all methods, we define a grid over salient hyper-parameters and tune over the validation split of \textit{mini}-ImageNet. To avoid a cumbersome per-dataset tuning, and evaluate model-agnosticity of methods, we then keep hyper-parameters fixed across all other experiments. 
        
        \magicpar{Architectures and checkpoints.} To provide the fairest comparison, all non-episodic methods are tuned and tested using off-the-shelf pre-trained checkpoints. All results except Fig. \ref{fig:barplots} are produced using the pre-trained ResNet-12 and Wide-ResNet 28-10 checkpoints provided by the authors from \cite{ye2020few}. As for episodically-finetuned models required by Snatcher \cite{jeong2021few} and FEAT \cite{ye2020few}, checkpoints are obtained from the authors' respective repositories. Finally, to challenge the model-agnosticity of our method, we resort to an additional set of 10 ImageNet pre-trained models covering three distinct architectures: ResNet-50 \cite{resnet} for CNNs, ViT-B/16 \cite{vit} for vision transformers, and Mixer-B/16 \cite{mlp_mixer} for MLP-Mixer. Most models used are taken from the excellent \textsc{TIMM} library \cite{rw2019timm}.

        \magicpar{Datasets and tasks.} We experiment with a total of 5 vision datasets. As standard FSC benchmarks, we use the \textit{mini}-ImageNet \cite{Vinyals16} dataset with 100 classes and the larger \textit{tiered}-ImageNet \cite{tiered_imagenet} dataset with 608 classes. We also experiment on more challenging cross-domain tasks formed by using 3 finer-grained datasets: the Caltech-UCSD Birds 200 \cite{cub} (CUB) dataset, with 200 classes, the FGVC-Aircraft dataset \cite{maji2013fine} with 100 classes, and the Fungi classification challenge \cite{schroeder2018fgvcx} with 1394 classes. Following standard FSOSR protocol, support sets contain $|\cseen|=5$ with 1 or 5 instances, or \textit{shots}, per class, and query sets are formed by sampling 15 instances per class, from a total of ten classes, the five closed-set classes and an additional set of $|\cunseen|=5$ open-set classes.

    \subsection{Results}
    % In what follow, we synthetize the important messages of this study, along with quantitative

\begin{figure}[t]
            \centering
            \includegraphics[width=\textwidth]{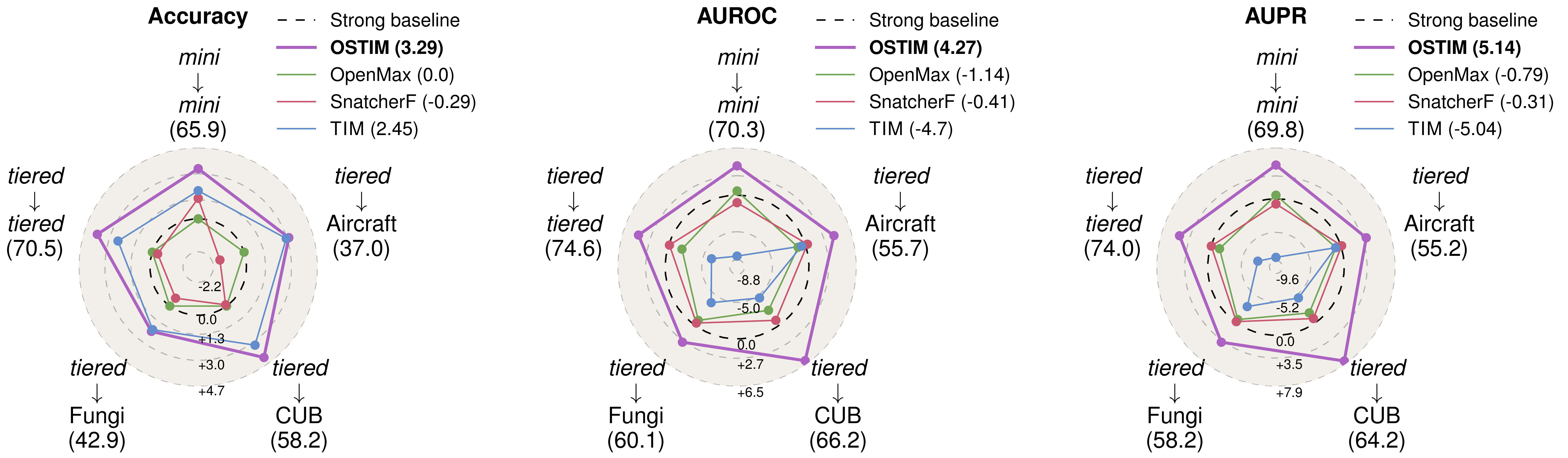} \\
            % \vspace{1em}
            % \includegraphics[width=\textwidth]{figures/spider_5.pdf}
            \caption{\textbf{Wider benchmarking.} Relative 1-shot performance of the best methods of each family w.r.t the \textit{Strong baseline}, across a set of 5 scenarios, including 3 with domain-shift. Each vertex represents one scenario, e.g. \textit{tiered}$\rightarrow$Fungi ($x$) means the feature extractor was pre-trained on \textit{tiered}-ImageNet, test tasks are sampled from Fungi, and the \textit{Strong Baseline} performance is $x$. For each method, the average relative improvement across the 5 scenarios is reported in parenthesis in the legend.}
            % Better viewed in colors and zoomed in.}
            \label{fig:spider_charts}
        \end{figure}

        \magicpar{Simplest inductive methods are competitive.} 
        A first surprising result comes from analyzing the performances of standard OOD detectors on the FSOSR problem. Table \ref{tab:benchmark_results_mini} shows that \knn~ and \textsc{PCA} outperform, by far, arguably more advanced methods that are OCVSM and Isolation Forest. This result contrasts with standard high-dimensional benchmarks \cite{zhao2019pyod} where \knn~ falls typically short of the latter. On the other hand, such results echo with Sec. \ref{subsec:difficulty_of_problem}, in which FSOSR was shown to introduce serious challenges that may expose advanced methods to overfit. In fact, Fig. \ref{fig:spider_charts} shows that across 5 scenarios, the combination SimpleShot \cite{wang2019simpleshot}+ \knn~ \cite{knn_detector} formed by the simplest FS-inductive classifier and the simplest inductive OOD detector is a strong baseline that outperforms all specialized open-set methods. We refer to this combination as \textit{Strong baseline} in Figures \ref{fig:spider_charts} and \ref{fig:barplots}. Additional results for the Wide-ResNet architecture are provided in the supplementary material.

        \magicpar{Transductive methods still improve accuracy but degrade outlier detection.} As shown in Table \ref{tab:benchmark_results_mini}, most transductive classifiers still offer a significant boost in closed-set accuracy, even in the presence of outliers in the query set. Note that this contrasts with findings from the semi-supervised literature, where standard methods drop below the baseline in the presence of even a small fraction of outliers \cite{yu2020multi,chen2020semi,saito2021openmatch, killamsetty2021retrieve}. We hypothesize that the deliberate under-parametrization of few-shot methods --typically only training a linear classifier--, required to avoid overfitting the support set, partly explains such robustness. However, transductive methods still largely underperform in OD, with AUROCs as low as 52 \% (50\% being a random detector) for LaplacianShot.

        \begin{wraptable}{r}{0.35\textwidth}
            \centering
            \small
            \vspace{-10pt}
            \caption{\textbf{OSTIM's ablation study} along three factors described in Sec. \ref{sec:ablation_study}. Results are produced on the 1-shot scenario on \textit{tiered}-ImageNet, with a ResNet-12. \\}
            \resizebox{.35\textwidth}{!}{
                \begin{tabular}{lcc}
                    \toprule
                    (i) Centering & Acc & AUPR \\
                    \midrule
                    No centering & 74.2 & 72.5 \\
                    $\mubf_{base}$ \cite{wang2019simpleshot} & 74.8 & 77.1 \\
                    $\mubf_{task}$ \cite{hu2021leveraging} & 74.8 & 78.7 \\
                    \toprule
                    (ii) Prototype refinement & Acc & AUPR \\
                    \midrule
                    At initialization & 71.6 & 77.5 \\
                    After 200 gradient steps & 74.8 & 78.7 \\
                    \toprule
                    (iii) Outlier prototype & Acc & AUPR \\
                    \midrule
                    Explicit dummy \cite{zhou2021learning} & 74.9 & 71.1 \\
                    Implicit Eq. \eqref{eq:outlier_logit} & 74.9 & 78.7 \\
                    \bottomrule
                \end{tabular}
            }
            \label{tab:ablation_study}
            \vspace{-12pt}
        \end{wraptable}  
    
        \magicpar{\OSTIM achieves the best trade-off.} Benchmark results in Table \ref{tab:benchmark_results_mini} show that \OSTIM competes with the best transductive methods in terms of closed-set accuracy, while consistently outperforming existing OOD or Open-Set competitors on outlier detection ability. Interestingly, while the gap between closed-set accuracy of transductive methods and inductive ones typically contracts with more shots, the outlier detection performance of \OSTIM remains largely superior to its inductive competitors even in the 5-shot scenario, where a consistent 6-7\% gap in AUROC and AUPR with the second-best method can be observed. We accumulate further evidence of \OSTIM's superiority by introducing 3 additional cross-domain scenarios in Fig. \ref{fig:spider_charts}, corresponding to a base model pre-trained on \textit{tiered}-ImageNet, but tested on CUB, Aircraft, and Fungi datasets. In such challenging scenarios, where both feature and class distributions shift, \OSTIM maintains consistent improvements, and even widens the gap in the \textit{tiered} $\rightarrow$ CUB setting, achieving a strong $7 \% +$ AUPR improvement over the Strong Baseline.

        \begin{figure}[t]
            \centering
            \includegraphics[width=0.9\textwidth]{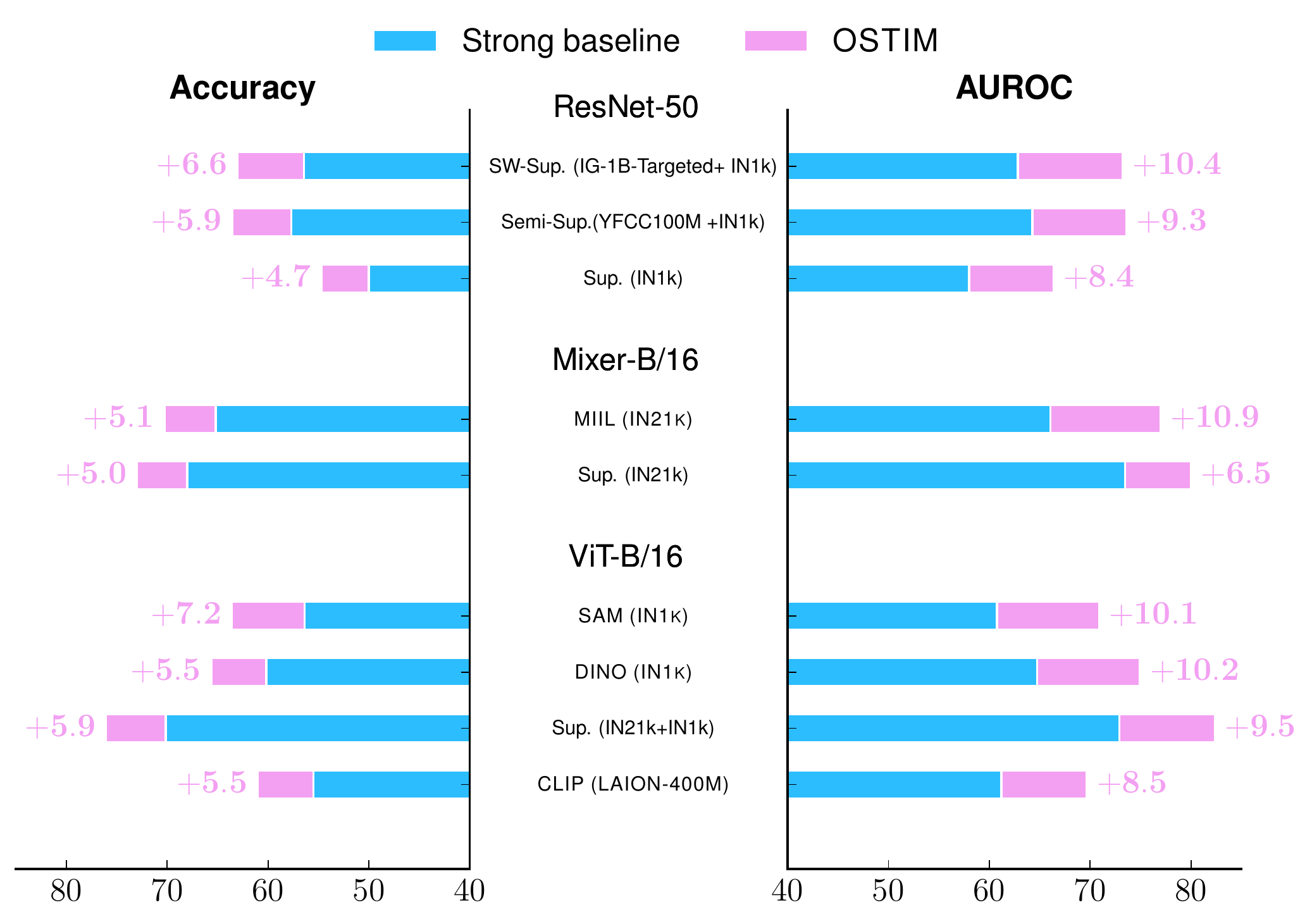}
            \caption{\textbf{Model agnosticity}. To evaluate this important quality, we compare \OSTIM to the Strong baseline on challenging 1-shot Fungi tasks. We experiment across 3 largely distinct architectures: ResNet-50 (CNN) \cite{resnet}, ViT-B/16 (Vision Transformer) \cite{vit}, and Mixer-B/16 (MLP-Mixer) \cite{mlp_mixer}. For each architecture, we include different types of pre-training, including Supervised (Sup.), Semi-Supervised, Semi-Weakly Supervised (SW Sup.) \cite{semi_sup}, \textsc{DINO} \cite{dino}, \textsc{SAM} \cite{sam}, \textsc{MIIL} \cite{miil}. Improvements over the baseline are consistently significant and generally higher than those observed with the ResNet-12 in Fig. \ref{fig:spider_charts}.}
            \label{fig:barplots}
        \end{figure}
        
        \magicpar{\OSTIM steps toward model-agnosticity.} 
        As a core motivation of this work, we evaluate OSTIM's \textit{model-agnosticity} by its ability to maintain consistent improvement over the \textit{Strong Baseline}, regardless of the model used, and without hyperparameter adjustment.  In that regard, we depart from the standard ResNet-12 and cover 3 largely distinct architectures, each encoding different inductive biases. To further strengthen our claim, for each architecture, we consider several training strategies spanning different paradigms -- unsupervised, supervised, semi and semi-weakly supervised -- and using different types of data --image, text--. Results in Fig. \ref{fig:barplots} show the relative improvement of \OSTIM w.r.t the strong baseline in the 1-shot scenario on the $\ast \rightarrow$ Fungi benchmark. Without any tuning, \OSTIM remains able to leverage the strong expressive power of large-scale models, and even consistently widens the gap with the strong baseline, achieving a remarkable performance of 76 \% accuracy and 82 \% AUROC with the ViT-B/16. We believe this set of results is very promising for two reasons. It first testifies how easy obtaining highly competitive results on difficult specialized tasks can be by combining \OSTIM with the latest models. Second, it legitimately leads to believe that as architectures and training strategies continue to improve, so will \OSTIM's performance.

    \magicpar{Ablation study.} \label{sec:ablation_study}
    We refer to the results in Table \ref{tab:ablation_study} to motivate the design choices of \OSTIM: (i) A task-specific instantiation of the centering vector $\mubf$ in Eq. \eqref{eq:inlier_logits} is important for outlier detection, but not strictly necessary as the base centering introduced in \cite{wang2019simpleshot}, already yields strong results when used along \OSTIM. (ii) Even at initialization, \OSTIM achieves high outlier detection performances but requires prototype refinement through the mutual information maximization in Eq. \eqref{eq:ottim_objective} to improve its CS accuracy. (iii) The inductive bias that consists in implicitly defining the \textit{outlier} prototype as the diametrically opposite of the average of support prototypes is crucial. Introducing and optimizing an independent prototype as in the large-scale open-set \textsc{PROSER} \cite{zhou2021learning} only adds up to the ambiguity of the few-shot problem, and ends up achieving poor outlier detection performances.

\section{Discussion and Limitations}
\label{sec:discussions}

In this study, we advocate for Transduction as a promising avenue to address the difficult FSOSR problem. Through the proposed implicit prototype idea, we show that the InfoMax method TIM can be successfully \textit{opened}. We further insist that the proposed technique does not necessitate any particular training process or model-specific parameter optimization, and can therefore be plugged effortlessly into the most expressive feature extractors. We hope that the promising results we obtained in this setting, using the latest advances in representation learning, will encourage the community to go beyond small residual networks and leverage these advances in our methods more often than we do today. Taking a step back, the opening technique we propose is in fact broadly applicable, and other transductive classification methods, \eg LaplacianShot or PT-MAP could be similarly \textit{opened}. In that regard, attempts at \textit{opening} other existing methods would broaden the scope of the current study. As a second limitation, we note that Transduction does not come for free. In particular, using unlabelled data makes transductive methods inherently sensitive to the statistical properties of the query set, \eg class balance or size. Such properties may greatly vary from one  practical situation to another. Consequently, an in-depth study of the impact of such properties on the performances of \OSTIM would surely help better identify the situations in which Transduction should be preferred over Induction, and vice-versa.

{\small
\bibliographystyle{ieee_fullname}
\bibliography{egbib}

\begin{thebibliography}{10}\itemsep=-1pt

\bibitem{antoniou2018train}
Antreas Antoniou, Harrison Edwards, and Amos Storkey.
\newblock How to train your maml.
\newblock {\em International Conference on Learning Representations (ICLR)},
  2019.

\bibitem{bendale2016towards}
Abhijit Bendale and Terrance~E Boult.
\newblock Towards open set deep networks.
\newblock In {\em Computer Vision and Pattern Recognition Conference (CVPR)},
  2016.

\bibitem{bennequin2019meta}
Etienne Bennequin.
\newblock Meta-learning algorithms for few-shot computer vision.
\newblock {\em arXiv}, 2019.

\bibitem{bennequin2021bridging}
Etienne Bennequin, Victor Bouvier, Myriam Tami, Antoine Toubhans, and
  C{\'e}line Hudelot.
\newblock Bridging few-shot learning and adaptation: New challenges of
  support-query shift.
\newblock {\em European Conference on Machine Learning and Principles and
  Practice of Knowledge Discovery in Databases (ECML-PKDD)}, 2021.

\bibitem{boudiaf2021few}
Malik Boudiaf, Hoel Kervadec, Ziko~Imtiaz Masud, Pablo Piantanida, Ismail
  Ben~Ayed, and Jose Dolz.
\newblock Few-shot segmentation without meta-learning: A good transductive
  inference is all you need?
\newblock In {\em Computer Vision and Pattern Recognition Conference (CVPR)},
  2021.

\bibitem{boudiaf2020transductive}
Malik Boudiaf, Ziko~Imtiaz Masud, J{\'e}r{\^o}me Rony, Jos{\'e} Dolz, Pablo
  Piantanida, and Ismail Ben~Ayed.
\newblock Transductive information maximization for few-shot learning.
\newblock In {\em Neural Information Processing Systems (NeurIPS)}, 2020.

\bibitem{dino}
Mathilde Caron, Hugo Touvron, Ishan Misra, Herv{\'e} J{\'e}gou, Julien Mairal,
  Piotr Bojanowski, and Armand Joulin.
\newblock Emerging properties in self-supervised vision transformers.
\newblock In {\em ICCV}, 2021.

\bibitem{Chen19}
Wei-Yu Chen, Yen-Cheng Liu, Zsolt Kira, Yu-Chiang Wang, and Jia-Bin Huang.
\newblock A closer look at few-shot classification.
\newblock In {\em International Conference on Learning Representations (ICLR)},
  2019.

\bibitem{sam}
Xiangning Chen, Cho-Jui Hsieh, and Boqing Gong.
\newblock When vision transformers outperform resnets without pre-training or
  strong data augmentations.
\newblock {\em International Conference on Learning Representations (ICLR)},
  2022.

\bibitem{chen2020semi}
Yanbei Chen, Xiatian Zhu, Wei Li, and Shaogang Gong.
\newblock Semi-supervised learning under class distribution mismatch.
\newblock In {\em Conference on Artificial Intelligence (AAAI)}, 2020.

\bibitem{dhillon2019baseline}
Guneet~Singh Dhillon, Pratik Chaudhari, Avinash Ravichandran, and Stefano
  Soatto.
\newblock A baseline for few-shot image classification.
\newblock In {\em International Conference on Learning Representations (ICLR)},
  2020.

\bibitem{vit}
Alexey Dosovitskiy, Lucas Beyer, Alexander Kolesnikov, Dirk Weissenborn,
  Xiaohua Zhai, Thomas Unterthiner, Mostafa Dehghani, Matthias Minderer, Georg
  Heigold, Sylvain Gelly, et~al.
\newblock An image is worth 16x16 words: Transformers for image recognition at
  scale.
\newblock {\em International Conference on Learning Representations (ICLR)},
  2021.

\bibitem{ge2017generative}
ZongYuan Ge, Sergey Demyanov, Zetao Chen, and Rahil Garnavi.
\newblock Generative openmax for multi-class open set classification.
\newblock {\em arXiv}, 2017.

\bibitem{goldblum2020unraveling}
Micah Goldblum, Steven Reich, Liam Fowl, Renkun Ni, Valeriia Cherepanova, and
  Tom Goldstein.
\newblock Unraveling meta-learning: Understanding feature representations for
  few-shot tasks.
\newblock In {\em International Conference on Machine Learning (ICML)}, 2020.

\bibitem{resnet}
Kaiming He, Xiangyu Zhang, Shaoqing Ren, and Jian Sun.
\newblock Deep residual learning for image recognition.
\newblock In {\em Computer Vision and Pattern Recognition Conference (CVPR)},
  2016.

\bibitem{hu2021leveraging}
Yuqing Hu, Vincent Gripon, and St{\'e}phane Pateux.
\newblock Leveraging the feature distribution in transfer-based few-shot
  learning.
\newblock In {\em International Conference on Artificial Neural Networks},
  2021.

\bibitem{jeong2021few}
Minki Jeong, Seokeon Choi, and Changick Kim.
\newblock Few-shot open-set recognition by transformation consistency.
\newblock In {\em Computer Vision and Pattern Recognition Conference (CVPR)},
  2021.

\bibitem{killamsetty2021retrieve}
Krishnateja Killamsetty, Xujiang Zhao, Feng Chen, and Rishabh Iyer.
\newblock Retrieve: Coreset selection for efficient and robust semi-supervised
  learning.
\newblock {\em Neural Information Processing Systems (NeurIPS)}, 34, 2021.

\bibitem{lazarou2021iterative}
Michalis Lazarou, Tania Stathaki, and Yannis Avrithis.
\newblock Iterative label cleaning for transductive and semi-supervised
  few-shot learning.
\newblock In {\em ICCV}, 2021.

\bibitem{copod_detector}
Zheng Li, Yue Zhao, Nicola Botta, Cezar Ionescu, and Xiyang Hu.
\newblock Copod: copula-based outlier detection.
\newblock In {\em International Conference on Data Mining (ICDM)}, 2020.

\bibitem{lichtenstein2020tafssl}
Moshe Lichtenstein, Prasanna Sattigeri, Rogerio Feris, Raja Giryes, and Leonid
  Karlinsky.
\newblock Tafssl: Task-adaptive feature sub-space learning for few-shot
  classification.
\newblock In {\em European Conference on Computer Vision (ECCV)}, 2020.

\bibitem{liu2020few}
Bo Liu, Hao Kang, Haoxiang Li, Gang Hua, and Nuno Vasconcelos.
\newblock Few-shot open-set recognition using meta-learning.
\newblock In {\em Computer Vision and Pattern Recognition Conference (CVPR)},
  2020.

\bibitem{iforest_detector}
Fei~Tony Liu, Kai~Ming Ting, and Zhi-Hua Zhou.
\newblock Isolation forest.
\newblock In {\em International Conference on Data Mining (ICMD)}, 2008.

\bibitem{liu2020prototype}
Jinlu Liu, Liang Song, and Yongqiang Qin.
\newblock Prototype rectification for few-shot learning.
\newblock In {\em European Conference on Computer Vision (ECCV)}, 2020.

\bibitem{maji2013fine}
Subhransu Maji, Esa Rahtu, Juho Kannala, Matthew Blaschko, and Andrea Vedaldi.
\newblock Fine-grained visual classification of aircraft.
\newblock {\em arXiv}, 2013.

\bibitem{neal2018open}
Lawrence Neal, Matthew Olson, Xiaoli Fern, Weng-Keen Wong, and Fuxin Li.
\newblock Open set learning with counterfactual images.
\newblock In {\em European Conference on Computer Vision (ECCV)}, 2018.

\bibitem{knn_detector}
Sridhar Ramaswamy, Rajeev Rastogi, and Kyuseok Shim.
\newblock Efficient algorithms for mining outliers from large data sets.
\newblock In {\em International Conference on Management of Data}, 2000.

\bibitem{tiered_imagenet}
Mengye Ren, Eleni Triantafillou, Sachin Ravi, Jake Snell, Kevin Swersky,
  Joshua~B Tenenbaum, Hugo Larochelle, and Richard~S Zemel.
\newblock Meta-learning for semi-supervised few-shot classification.
\newblock {\em International Conference on Learning Representations (ICLR)},
  2018.

\bibitem{miil}
Tal Ridnik, Emanuel Ben-Baruch, Asaf Noy, and Lihi Zelnik-Manor.
\newblock Imagenet-21k pretraining for the masses.
\newblock {\em Neural Information Processing Systems (NeurIPS)}, 2021.

\bibitem{saito2021openmatch}
Kuniaki Saito, Donghyun Kim, and Kate Saenko.
\newblock Openmatch: Open-set semi-supervised learning with open-set
  consistency regularization.
\newblock {\em Neural Information Processing Systems (NeurIPS)}, 2021.

\bibitem{scheirer2012toward}
Walter~J Scheirer, Anderson de Rezende~Rocha, Archana Sapkota, and Terrance~E
  Boult.
\newblock Toward open set recognition.
\newblock {\em PAMI}, 2012.

\bibitem{ocsvm_detector}
Bernhard Sch{\"o}lkopf, John~C Platt, John Shawe-Taylor, Alex~J Smola, and
  Robert~C Williamson.
\newblock Estimating the support of a high-dimensional distribution.
\newblock {\em Neural computation}, 2001.

\bibitem{schroeder2018fgvcx}
Brigit Schroeder and Yin Cui.
\newblock Fgvcx fungi classification challenge 2018.
\newblock 2018.

\bibitem{pca_detector}
Mei-Ling Shyu, Shu-Ching Chen, Kanoksri Sarinnapakorn, and LiWu Chang.
\newblock A novel anomaly detection scheme based on principal component
  classifier.
\newblock Technical report, 2003.

\bibitem{snell2017prototypical}
Jake Snell, Kevin Swersky, and Richard~S Zemel.
\newblock Prototypical networks for few-shot learning.
\newblock {\em Neural Information Processing Systems (NeurIPS)}, 2017.

\bibitem{mlp_mixer}
Ilya~O Tolstikhin, Neil Houlsby, Alexander Kolesnikov, Lucas Beyer, Xiaohua
  Zhai, Thomas Unterthiner, Jessica Yung, Andreas Steiner, Daniel Keysers,
  Jakob Uszkoreit, et~al.
\newblock Mlp-mixer: An all-mlp architecture for vision.
\newblock {\em Neural Information Processing Systems (NeurIPS)}, 2021.

\bibitem{vaze2021open}
Sagar Vaze, Kai Han, Andrea Vedaldi, and Andrew Zisserman.
\newblock Open-set recognition: A good closed-set classifier is all you need.
\newblock In {\em International Conference on Learning Representations (ICLR)},
  2022.

\bibitem{veilleux2021realistic}
Olivier Veilleux, Malik Boudiaf, Pablo Piantanida, and Ismail Ben~Ayed.
\newblock Realistic evaluation of transductive few-shot learning.
\newblock {\em Neural Information Processing Systems (NeurIPS)}, 2021.

\bibitem{Vinyals16}
Oriol Vinyals, Charles Blundell, Timothy Lillicrap, Koray Kavukcuoglu, and Daan
  Wierstra.
\newblock Matching networks for one shot learning.
\newblock In {\em Neural Information Processing Systems (NeurIPS)}, 2016.

\bibitem{wang2019simpleshot}
Yan Wang, Wei-Lun Chao, Kilian~Q Weinberger, and Laurens van~der Maaten.
\newblock Simpleshot: Revisiting nearest-neighbor classification for few-shot
  learning.
\newblock {\em arXiv}, 2019.

\bibitem{wang2020instance}
Yikai Wang, Chengming Xu, Chen Liu, Li Zhang, and Yanwei Fu.
\newblock Instance credibility inference for few-shot learning.
\newblock In {\em Computer Vision and Pattern Recognition Conference (CVPR)},
  2020.

\bibitem{cub}
Peter Welinder, Steve Branson, Takeshi Mita, Catherine Wah, Florian Schroff,
  Serge Belongie, and Pietro Perona.
\newblock Caltech-ucsd birds 200.
\newblock 2010.

\bibitem{rw2019timm}
Ross Wightman.
\newblock Pytorch image models.
\newblock \url{https://github.com/rwightman/pytorch-image-models}, 2019.

\bibitem{semi_sup}
I~Zeki Yalniz, Herv{\'e} J{\'e}gou, Kan Chen, Manohar Paluri, and Dhruv
  Mahajan.
\newblock Billion-scale semi-supervised learning for image classification.
\newblock {\em arXiv}, 2019.

\bibitem{ye2020few}
Han-Jia Ye, Hexiang Hu, De-Chuan Zhan, and Fei Sha.
\newblock Few-shot learning via embedding adaptation with set-to-set functions.
\newblock In {\em Computer Vision and Pattern Recognition Conference (CVPR)},
  2020.

\bibitem{yu2020multi}
Qing Yu, Daiki Ikami, Go Irie, and Kiyoharu Aizawa.
\newblock Multi-task curriculum framework for open-set semi-supervised
  learning.
\newblock In {\em European Conference on Computer Vision (ECCV)}, 2020.

\bibitem{zhao2019pyod}
Yue Zhao, Zain Nasrullah, and Zheng Li.
\newblock Pyod: A python toolbox for scalable outlier detection.
\newblock {\em JMLR}, 2019.

\bibitem{zhou2021learning}
Da-Wei Zhou, Han-Jia Ye, and De-Chuan Zhan.
\newblock Learning placeholders for open-set recognition.
\newblock In {\em Computer Vision and Pattern Recognition Conference (CVPR)},
  2021.

\bibitem{ziko2020laplacian}
Imtiaz Ziko, Jose Dolz, Eric Granger, and Ismail~Ben Ayed.
\newblock Laplacian regularized few-shot learning.
\newblock In {\em International Conference on Machine Learning (ICML)}, 2020.

\end{thebibliography}
}

\clearpage

\clearpage
\appendix
\appendix
\section{Metrics}

Here we provide some details about the metrics used in Section \ref{sec:experiments}

    \textbf{Acc}: the classification accuracy on the closed-set instances of the query set (\ie $y^q \in \mathbb{C_S}$).
    
    \textbf{AUROC}: the area under the ROC curve is an almost mandatory metric for any OOD detection task. For a set of outlier predictions in $[O,1]$ and their ground truth ($0$ for inliers, $1$ for outliers), any threshold $\gamma \in [O,1]$ gives a true positive rate $\textit{TP}(\gamma)$ (\ie recall) and a false positive rate $\textit{FP}(\gamma)$. By rolling this threshold, we obtain a plot of \textit{TP} as a function of \textit{FP} \ie the ROC curve. The area under this curve is a measure of the discrimination ability of the outlier detector. Random predictions lead to an AUROC of $50\%$.
    
    \textbf{AUPR}: the area under the precision-recall (PR) curve is also a common metric in OOD detection. With the same principle as the ROC curve, the PR curve plots the precision as a function of the recall. Random predictions lead to an AUPR equal to the proportion of outliers in the query set \ie $50\%$ in our set-up.
    
    \textbf{Prec@0.9}: the precision at $90\%$ recall is the achievable precision on the few-shot open-set recognition task when setting the threshold allowing a recall of $90\%$ for the same task. While AUROC and AUPR are global metrics, \textit{Prec@0.9} measures the ability of the detector to solve a specific problem, which is the detection of almost all outliers (\eg for raising an alert when open-set instances appear so a human operator can create appropriate new classes). Since all detectors are able to achieve high recall with a sufficiently permissive threshold $\gamma$, an excellent way to compare them is to measure the precision of the predictor at a given level of recall (\ie the proportion of false alarms that the human operator will have to handle). Random predictions lead to a \textit{Prec@0.9} equal to the proportion of outliers in the query set \ie $50\%$ in our set-up.

\section{Additional results}

\begin{figure}
        \centering
        \includegraphics[width=0.4\textwidth]{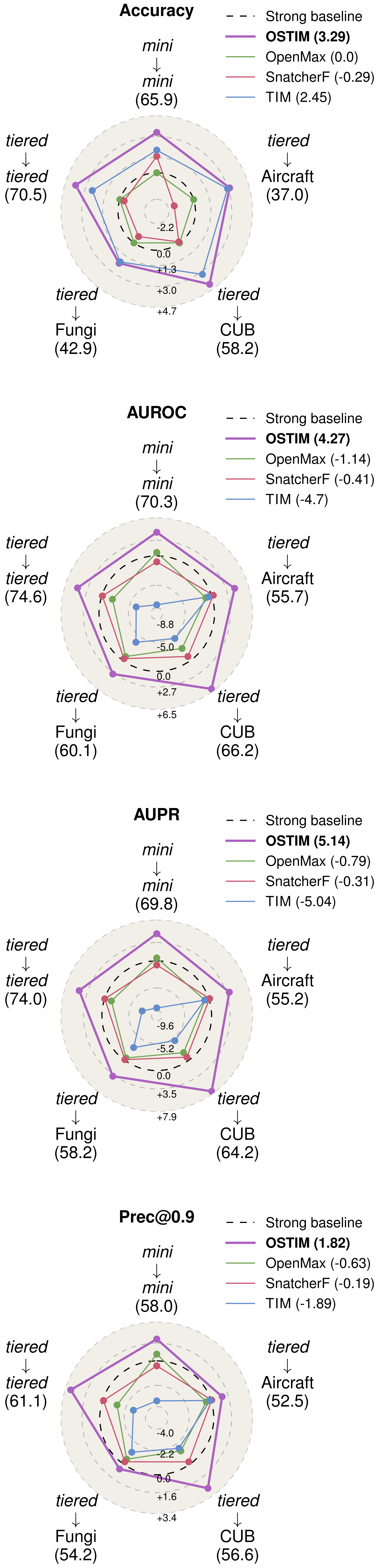}
        \includegraphics[width=0.4\textwidth]{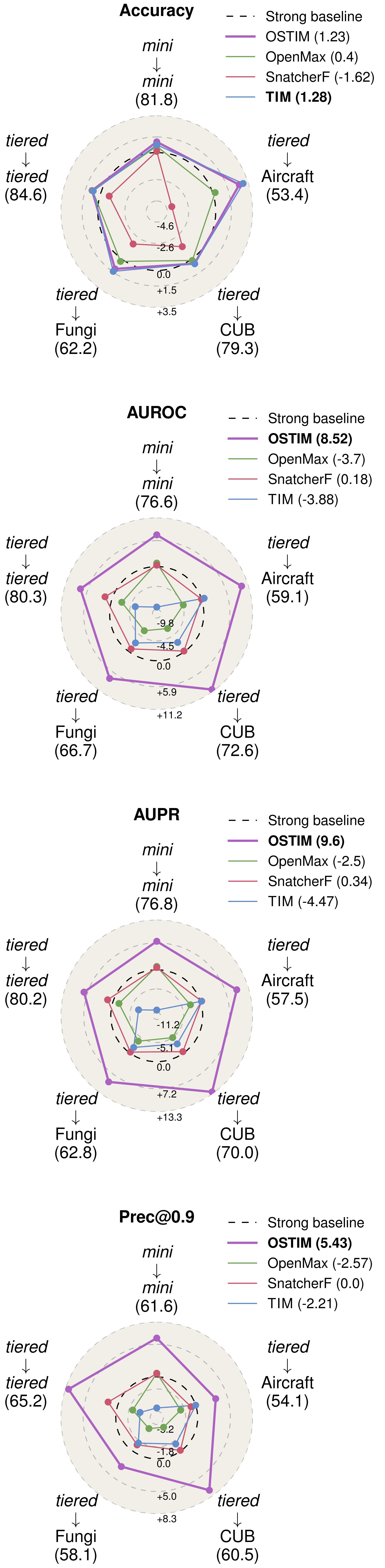}
        \caption{Complete version of Fig. \ref{fig:spider_charts} with a ResNet-12. (Left column): 1-shot. (Right column): 5-shot.}
        \label{fig:full_spider_resnet12}
    \end{figure}

\begin{figure}
        \centering
        \includegraphics[width=0.4\textwidth]{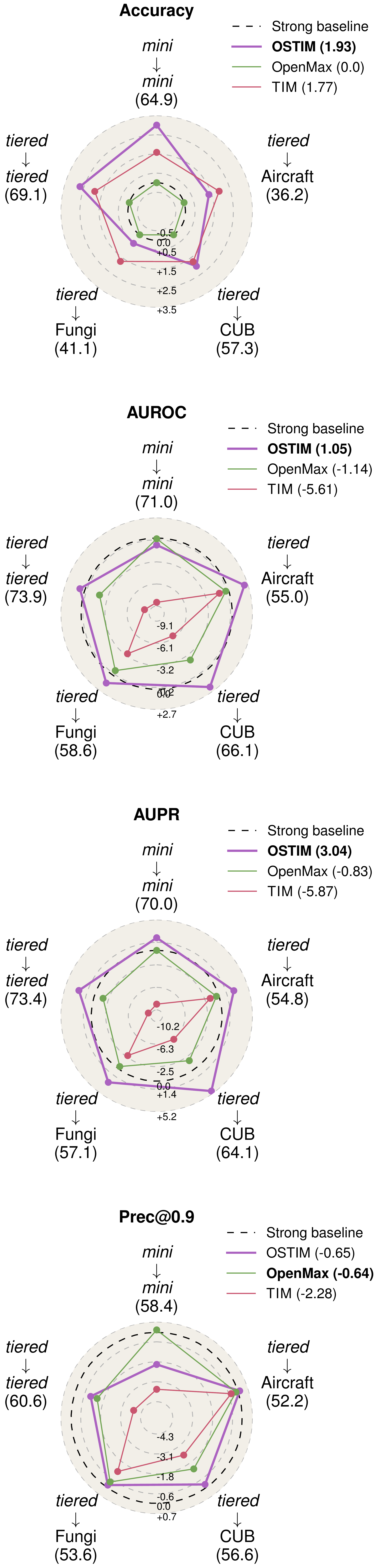}
        \includegraphics[width=0.4\textwidth]{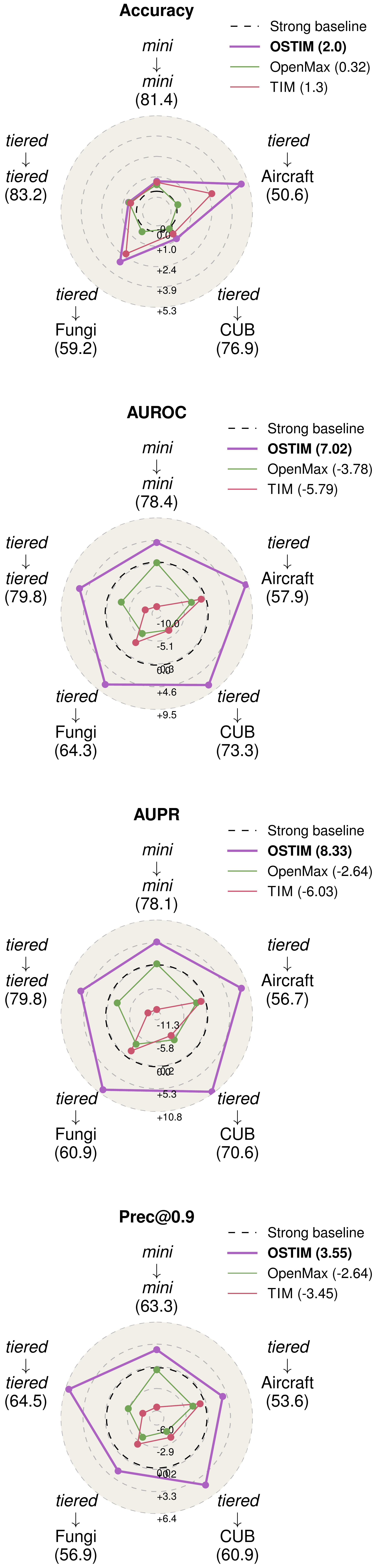}
        \caption{Complete version of Fig. \ref{fig:spider_charts} with a WideResNet 28-10. (Left column): 1-shot. (Right column): 5-shot. SnatcherF was not included in this plot because a yet misdiagnosed problem occurred with the provided \textit{tiered}-ImageNet checkpoint.}
        \label{fig:full_spider_wrn}
    \end{figure}

    Table \ref{tab:benchmark_results_tiered} shows the benchmark results on \textit{tiered}-ImageNet, and exhibits the same trends observed on \textit{mini}-ImageNet in Section \ref{sec:experiments}. Furthermore, we provide a more complete version of Fig. \ref{fig:spider_charts} in Fig. \ref{fig:full_spider_resnet12} and \ref{fig:full_spider_wrn}, showing the additional Prec@0.9 metric, along with the results on the WRN2810 provided by \cite{ye2020few}.
    
    \begin{table}[t]
        \centering
        \small
        \caption{\textbf{Standard Benchmarking}. Evaluating different families of methods on the Few-Shot Open-Set problem, on the popular \textit{tiered}-ImageNet, using a ResNet-12. For each column, a light-gray standard deviation is indicated, corresponding to the maximum deviation observed across methods for that metric. Best methods are shown in bold. Results for PEELER$^\star$ are reported from \cite{jeong2021few}.}
        \resizebox{\textwidth}{!}{
            \begin{tabular}{lccccccccc}
                \multicolumn{10}{c}{\large\textbf{\textit{tiered}-ImageNet}} \vspace{0.5em}\\
                \toprule
                \multirow{3}{*}{Strategy} & \multirow{3}{*}{Method} & \multicolumn{4}{c}{1-shot} & \multicolumn{4}{c}{5-shot} \\
                \cmidrule(lr){3-6} \cmidrule(lr){7-10}
                & & Acc & AUROC & AUPR & Prec@0.9 & Acc & AUROC & AUPR & Prec@0.9 \\
                & & \std{0.74} & \std{0.76} & \std{0.71} & \std{0.52} & \std{0.52} & \std{0.68} & \std{0.75} & \std{0.57} \\
                \midrule
                \multirow{6}{*}{OOD detection} & \knn \cite{knn_detector} & - & 74.62 & 73.99 & 61.1 & - & 80.32 & 80.15 & 65.24 \\
                                               & IForest \cite{iforest_detector} & - & 55.03 & 54.56 & 51.91 & - & 62.46 & 61.32 & 54.53 \\
                                               & OCVSM \cite{ocsvm_detector} & - & 71.72 & 71.98 & 58.68 & - &  70.85 & 67.93 & 60.88 \\
                                               & PCA \cite{pca_detector} & - & 68.78 & 67.74 & 57.68 & - & 76.37 & 76.55 & 61.5 \\
                                               & COPOD \cite{copod_detector} & - & 50.99 & 52.05 & 51.1 & - & 52.53 & 53.32 & 51.34 \\
                                               & HBOS  & - &  57.77 & 57.0 & 53.1 & - & 61.06 & 60.02 & 54.07 \\

                \toprule
                \multirow{3}{*}{Inductive classifiers} & SimpleShot \cite{wang2019simpleshot} & 70.52 & 70.39 & 68.42 & 58.99 & 84.65 & 77.67 & 76.24 & 63.5 \\
                                                       & Baseline ++ \cite{Chen19} & 70.53 & 70.34 & 68.32 & 59.03 & 84.78 & 74.01 & 72.47 & 61.25 \\
                                                       & FEAT \cite{ye2020few} & 70.15 & 52.43 & 56.44 & 50.0 & 83.79 & 53.31 & 59.81 & 50.0 \\
                \midrule
                \multirow{4}{*}{Inductive Open-Set} & PEELER$^\star$ \cite{liu2020few} & 69.51 & 65.20 & - & - & 84.10 & 73.27 & - & - \\
                                                    & SnatcherF \cite{jeong2021few}  & 70.15 & 74.51 & 73.94 & 61.01 & 83.79 & 81.97 & 81.65 & 66.78  \\
                                                    & OpenMax \cite{bendale2016towards} & 70.52 & 72.71 & 72.6 & 59.75 & \textbf{85.44} & 77.94 & 78.48 & 62.86 \\
                                                    & PROSER \cite{zhou2021learning} & 68.96 & 70.61 & 70.73 & 57.99 & 82.87 & 75.8 & 76.66 & 60.71  \\
                \toprule
                \multirow{5}{*}{Transductive classifiers} & LaplacianShot \cite{ziko2020laplacian} & \textbf{76.19} & 58.39 &     58.69 &     53.96 & \textbf{85.77} & 63.66 &    63.61 &     55.11 \\
                                                          & BDCSPN \cite{liu2020prototype} &  74.80 & 62.58 &    62.23 &    54.92 & \textbf{85.30}  & 67.43 &     67.49 &     56.25 \\
                                                          & TIM-GD \cite{boudiaf2020transductive} & 72.89 & 68.46 &     66.37 &     58.24 & \textbf{85.38} & 74.71 &     73.02 &     61.69 \\
                                                          & PT-MAP \cite{hu2021leveraging} & 71.39 & 64.86 &    63.39 &     56.57 &  82.66 & 71.08 &    69.65 & 59.14 \\
                                                          & LR-ICI \cite{wang2020instance} & 74.18 & 45.04 & 48.73 & 49.85 & 84.27 & 45.66 &  50.02 & 49.98 \\
                \midrule
                \rowcolor{lightsalmon!25} Transductive Open-Set & \OSTIM (ours) & 74.32 & \textbf{79.00} &  \textbf{79.11} & \textbf{64.04} & \textbf{85.50}  & \textbf{87.87} & \textbf{88.24} &  \textbf{73.08} \\
                \bottomrule
            \end{tabular}
        }
        \label{tab:benchmark_results_tiered}
    \end{table}

%%%%%%%%%%%%%%%%%%%%%%%%%%%%%%%%%%%%%%%%%%%%%%%%%%%%%%%%%%%%

\end{document}